\documentclass{article}

\PassOptionsToPackage{numbers, compress}{natbib}
\usepackage[final]{nips_2018}
  
\usepackage[utf8]{inputenc} 
\usepackage[T1]{fontenc}    
\usepackage{hyperref}       
\usepackage{url}            
\usepackage{booktabs}       
\usepackage{amsfonts}       
\usepackage{nicefrac}       
\usepackage{microtype}      
 
\usepackage{natbib}
\usepackage{amsmath}
\usepackage{setspace}
\usepackage{mathtools}
\usepackage{accents}
\usepackage{amssymb}
\usepackage{enumitem}
\usepackage{hyperref}
\usepackage{enumitem}
\usepackage{longtable}
\usepackage{fixmath}
\usepackage{amsmath}
\usepackage{wrapfig}
\usepackage{color}
\usepackage{algorithm,algpseudocode}
\usepackage{comment}
\usepackage{subfig}
\usepackage{soul}

\usepackage{epsfig}
\usepackage{multirow}
\usepackage{threeparttable}

\title{Combined Convolutional and Recurrent Neural Networks for Hierarchical Classification of Images}

\author{
  Jaehoon Koo\\
  Department of Industrial Engineering and Management Sciences\\
  Northwestern University\\
  Evanston, IL 60208 \\
  \texttt{jaehoonkoo2018@u.northwestern.edu} \\
   \And
   Diego Klabjan \\
  Department of Industrial Engineering and Management Sciences\\
  Northwestern University\\
  Evanston, IL 60208 \\
  \texttt{d-klabjan@northwestern.edu} \\
   \And
   Jean Utke \\
  Analytics Center of Excellence \\
  Allstate\\
  Northbrook, IL 60062 \\
  \texttt{jutke@allstate.com} \\
}

\begin{document}

\maketitle
\begin{abstract}
Deep learning models based on CNNs are predominantly used in image classification tasks. Such approaches, assuming independence of object categories, normally use a CNN as a feature learner and apply a flat classifier on top of it. Object classes in many settings have known hierarchical relations, and classifiers exploiting these relations should perform better. We propose hierarchical classification models combining a CNN to extract hierarchical representations of images, and an RNN or sequence-to-sequence model to capture a hierarchical tree of classes. In addition, we apply residual learning to the RNN part in order to facilitate training our compound model and improve generalization of the model. Experimental results on a public and a real world proprietary dataset of images show that our hierarchical networks perform better than state-of-the-art CNNs. 
\end{abstract}

\section{Introduction}

In computer vision, allocating labels to images is a fundamental problem, and it serves as a building block for various image recognition tasks such as image localization, object detection, and scene parsing \cite{Hu2016}. Over the past years, deep learning methods have made tremendous progress in these classification tasks. Especially, many approaches based on convolutional neural networks (CNNs) made significant advances in large-scale image classification \cite{He2015,Hu2018,Krizhevsky2012,Simonyan2014,Woo2018}. It is common to assume that separability of object categories is pronounced \cite{Yan2015}, and a multi-class or binary classifier is selected to label images \cite{Hu2016}. 

\begin{figure}[t]
\begin{center}
\includegraphics[width=0.7\textwidth]{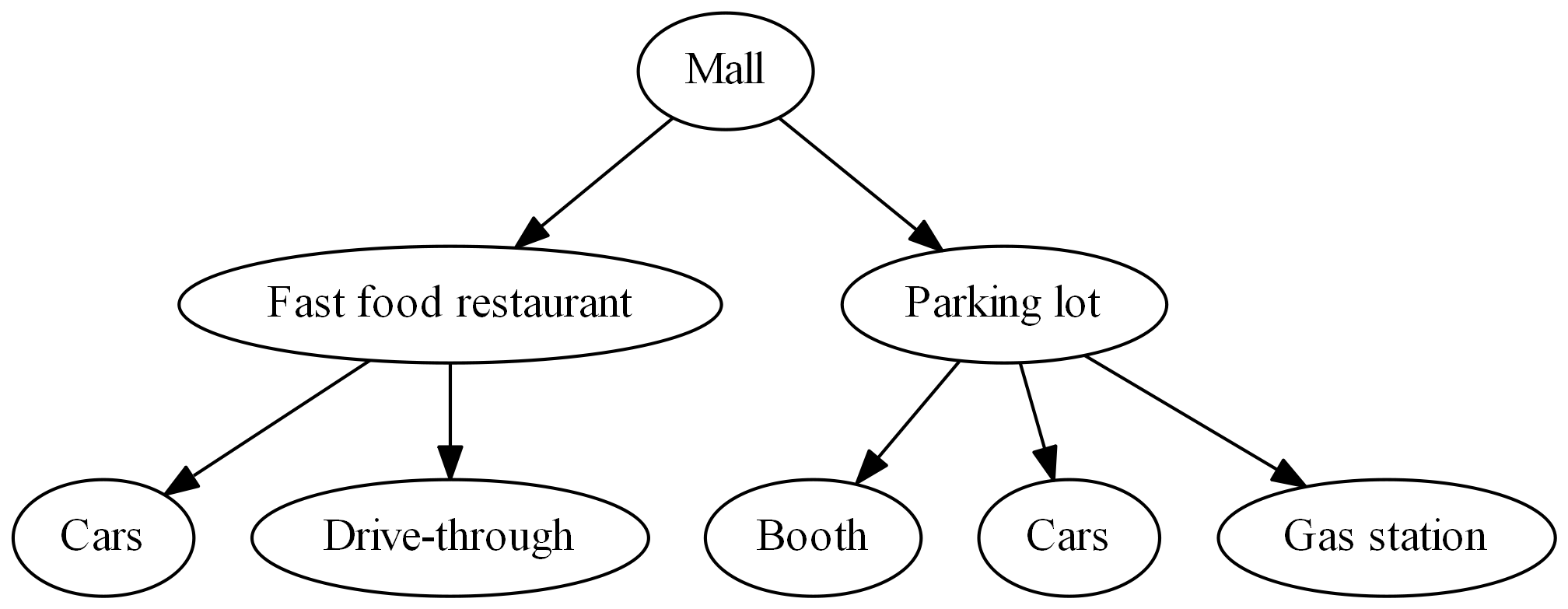}
\end{center}
   \caption{An example of a class tree}
 \label{hc-eg-1}
\end{figure}

Object categories in some settings are related to each other by means of a taxonomy. This phenomenon is typically present in datasets with a large number of categories \cite{Yan2015}. The categories of images can be represented by a tree based on two types of hierarchies: 1) Has-A hierarchy is present when each parent node physically contains some parts of each child node, and 2) Is-A hierarchy is exhibited when a parent node semantically contains child nodes, i.e. a child object is a type of the parent object. Models that can exploit details of objects lead to better classification performance. In some object classification tasks, objects contain detailed objects, and reliably classifying high level objects lead to classifying detailed objects correctly. In such settings, we can build a Has-A hierarchical tree of categories, and models that can capture hierarchical relationships are required. Consider an investment or commercial real estate firm relying on satellite images of malls to, for example, gauge investments. We have objects of `Booth,' `Cars,' and `Gas station' contained in images of parking lots. These categories have Has-A relationships, and a hierarchical tree of classes can be built as shown in Figure \ref{hc-eg-1}. To classify an image of `Booth,' we need a model to find a path of `Mall'-`Parking lot'-`Booth.' In this work, we consider the classification problem where we are given a tree of classes, and for an image we need to assign `a path' in the tree.


\begin{figure*}
\begin{center}
\includegraphics[width=0.8\textwidth]{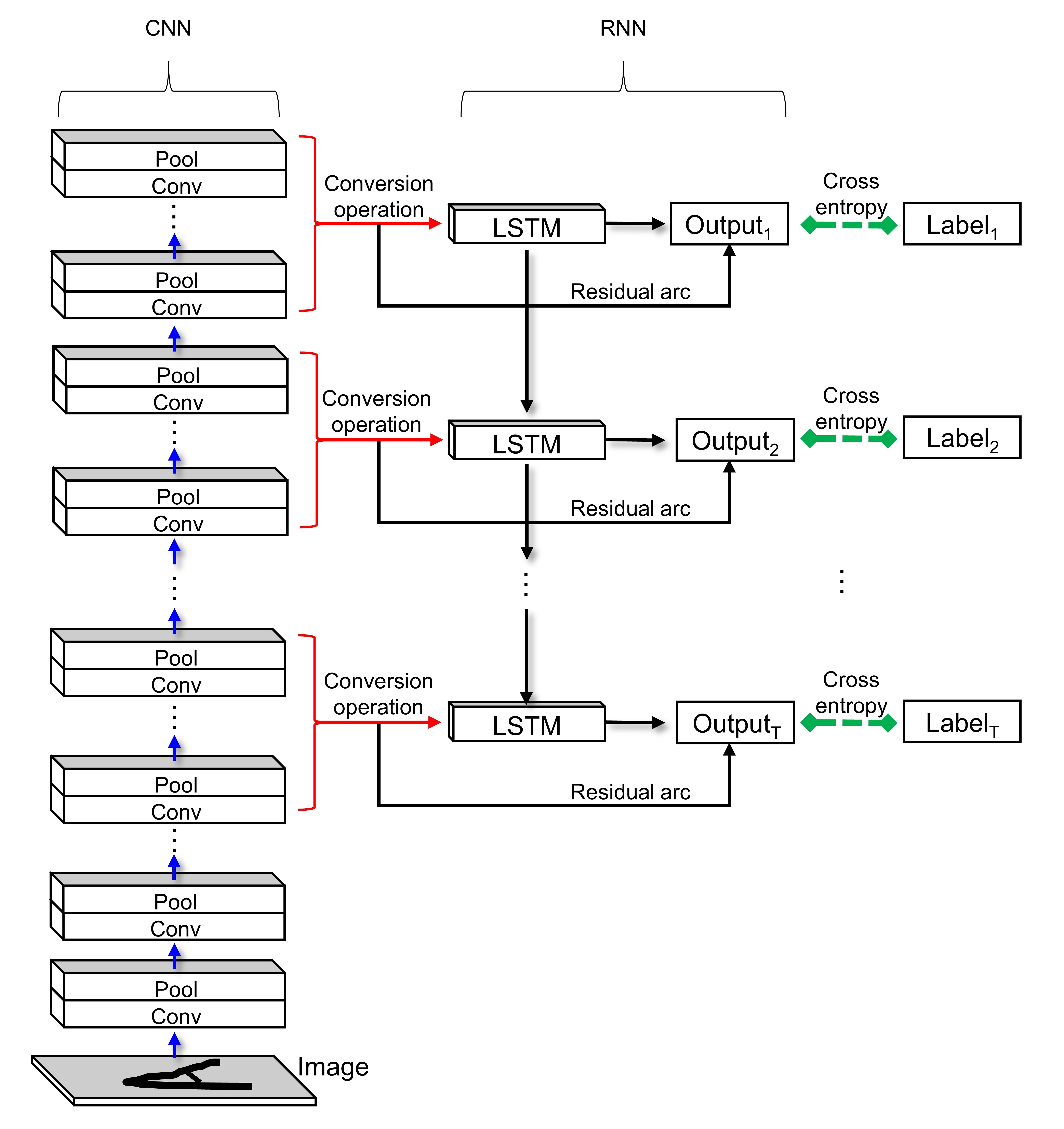}
\end{center}
   \caption{The model}
\label{hc1}
\end{figure*}

We propose hierarchical classification models for images, named \textit{deep hierarchical neural networks}, that extract hierarchical representations of images from a CNN and, by using a recurrent neural network (RNN), find a label path in the hierarchical class tree to predict labels of an image. Recent studies reveal that CNN features learn hierarchical representations of images at different layers representing an image ranging from detailed, part-level, to abstract, object-level \cite{Yosinski2015}. Part-level representations are typically captured at lower layers of the CNN, and object-level representations are learned at higher layers. Because of  this insight, it is conceivable to associate the different level feature maps with the different depth layers in the hierarchical label tree. High level features should be able to classify top layers in the tree while low level features focusing on details are suitable to predict classes in the bottom layers of the tree. It is natural to view a path in the tree as a sequence and then to model it via an RNN. For these reasons, we combine an RNN or sequence-to-sequence network (S2S) to classify a target sequence with a CNN. As a result we predict target paths rather than a single label. The proposed networks consist of three parts: 1) a CNN takes a raw image as input, and produces convolutional features at each layer, 2) the features at different layers of the CNN are converted to a vector of fixed dimension, and 3) an RNN or S2S takes the converted CNN features as input, and outputs predictions at each level of the label tree. Figure \ref{hc1} presents the structure of the proposed models.

To facilitate training of our compound model, we apply an alternating training scheme between the CNN and RNN sub models. Under this scheme, we alternate updating one while keeping the other frozen in the beginning of training and then unfreeze the entire network in the final phase of training. Such a scheme is needed because each sub model pursues different learning purposes in that the CNN learns representations of images and the RNN learns sequential behaviors of the classes. Alternating prevents both learning tasks from diverging in the early stage of training, consequently leading to better classification performance. In addition, different methods such as a linear, convolutional, and pooling operation are used to coerce the varying dimensions of the CNN features to the fixed dimension vector the RNN takes as input. The pooling retains much of the spatial information of the trained CNN features and avoids additional trainable parameters.

In our study, we use a real world, proprietary dataset of images from the insurance industry and a public dataset, Open Images \cite{OpenImages2}. Categories of both datasets have mainly Has-A relationships. We compare our models to state-of-the-art CNNs, and find that our models perform better. We conclude that our models can learn a hierarchical tree with both fixed- and variable-length target paths.


The main contributions of this work are as follows.
\begin{enumerate}
 \item We suggest a new structure of deep neural networks for hierarchical classification of images. Our models extract features from different CNN layers, and feed them to an RNN or S2S to learn a hierarchical path of categories. Our models can learn both fixed- and variable-length target paths; CNN-RNN are for fixed and CNN-S2S are for variable path lengths. 
 \item We apply residual learning to the RNN part in order to facilitate training of our compound model and improve generalization of the model.
\end{enumerate}

The rest of this paper is organized as follows. In Section \ref{sec:LR}, the related literature is discussed. Section \ref{sec:models} describes the proposed models, and Section \ref{sec:exp} provides a computational study including experimental details and analysis of the experimental results. Conclusions are given in Section \ref{sec:conclusion}.


\section{Related Work} \label{sec:LR}

Hierarchical structures have been studied for image recognition by using standard computer vision \cite{Tousch2012}. Related literature is categorized based on how a hierarchy is constructed \cite{Yan2015}; a hierarchy is predefined in \cite{Deng2012,Jia2013,Marszalek2007,Verma2012}, and it is trained by top-down and bottom-up methods in \cite{Bannour2012,Deng2011,Li2010,Liu2013,Marszalek2008,Salakhutdinov2011,Sivic2008}.

In the past, researchers adapted CNNs to hierarchical classification. Srivastava et al.~\cite{Srivastava2013} introduce CNNs to hierarchical classification. Their proposed method improves the performance of minority classes over standard CNN by incorporating priors imposed by a tree structure of the classes. Xiao et al.~\cite{Xiao2014} suggest CNN based hierarchical networks; each branch model predicts a super-class, and leaf models return final predictions. Yan et al.~\cite{Yan2015} suggest a hierarchical deep neural net that embeds CNNs into a two-level hierarchy of easy and difficult classes where the hierarchy is built automatically. The model uses coarse category classifiers for easy classes, and fine category classifiers for difficult classes. Schwing and Urtasun~\cite{Schwing2015} proposed a method for hierarchical semantic segmentation. They combine a Markov random field model that is used for segmentation with a CNN to extract image representations. All these works rely on using CNNs in their models to obtain a better feature learner for images while we approach the problem from the perspective of improving prediction of target label paths by combining an RNN or S2S with CNNs.


Approaches combining CNNs and RNNs have been studied to solve different image classification tasks such as scene parsing, object detection, image captioning, etc. Such CNN-RNN frameworks use the final feature map from the CNN and use it as an input to RNN (possibly combined with other features such as caption). Such a network takes advantages of the CNN's representational feature learning over images and the RNN's high performance in capturing sequential information. Deng et al.~\cite{Deng2016} propose an RNN that trains a graph structure for recognition of group activities. Stewart et al. \cite{Stewart2016} propose a model for object detection. The proposed model combines a CNN that encodes an image into features with an LSTM that decodes the encoded information into a set of people detections. In \cite{Liang2015}, a CNN-RNN model for object recognition is suggested by incorporating recurrent connections into each convolutional layer. It improves capturing of context information, which is important for object recognition. Wang et al.~\cite{Wang2016} propose a CNN-RNN framework for multi-label image classification. The proposed model produces class probabilities by concatenating CNN features and outputs of an RNN that takes a label vector as input. Shi et al.~\cite{Shi2015} propose a convolutional Long Short-Term Memory (ConvLSTM) in which convolutional operations are embedded in every LSTM layer. They show that ConvLSTM captures spatiotemporal correlations. Guo et al.~\cite{Guo2018} suggest several models to classify coarse- and fine-level categories of a semantic hierarchy; one of their models combines CNN and RNN so that top CNN features are input to RNN. 
Our approach is different from these methods since we exploit CNN features at each layer rather than only at the top layer. ConvLSTM overlays an RNN to each layer however its purpose is completely different; it does not focus on hierarchical classes but rather on sequences of images.

Recent papers suggest methods that consider CNN features from different layers, not only from the top layer, for hierarchical classification. Zhu and Bain~\cite{Zhu2017} suggest methods that take features at different middle layers of a CNN for coarse classes, and those at the top CNN layer for fine classes. Their network does not correlate the extracted CNN features to the final prediction; the extracted CNN features are trained independently without considering them as a sequence. Wehrmann et al.~\cite{Wehrmann2018} also propose a method considering features from middle layers of deep neural networks. They introduced an RNN to fit a hierarchical tree by inputting the extracted  features of a feed forward network. Their network feeds raw inputs from each layer to RNN (which is possible if all layers have the same number of neurons; not the case in CNNs) and its performance on image recognition tasks based on CNN is not studied. They also do not introduce the notion of residual arcs which we find to be  of great importance and they do not consider a S2S setting which is required if paths in the tree are of different length.

\section{Proposed models} \label{sec:models}

In this section, we describe the proposed models that predict target paths in a hierarchical class tree. Our models extract hierarchical features from a CNN taking an image as input, and feed the extracted features to an RNN if tree paths have the same lengths. The RNN part is replaced by S2S if tree paths have variable lengths. For this reason, we present two models, CNN-RNN and CNN-S2S.

\subsection{Fixed path length tree model: CNN-RNN} \label{sec:FPL}

We propose a hierarchical fixed path length (FPL) classification model to fit a class tree that has target paths with a fixed-length. To this end, we are given a rooted tree $R$ where each node corresponds to a class. We assume that each leaf node is of the same depth $T+1$. The root node corresponds to an artificial class. A training sample consists of $(x,y)$ where $x$ is an image and $y$ is a path from the root node to a leaf in $R$. By our assumption on $R$, every $y$ has the sample number $T$ of labels. In this model, an RNN is combined with a CNN. Our model starts with a CNN, a feature learner, that extracts hierarchical features representing part-level and object-level of images. A CNN is used since CNN features learn spatial representations of images through its local-connectivity of the networks, i.e. the features are learned locally; and the extracted features at different layers have hierarchical relations \cite{Yosinski2015,Zeiler2014}. In order to feed features from different CNN layers to the RNN, a conversion process is required. The dimensions of CNN features at each layer are different for combinations of convolution and pooling layers. However, RNN input dimensions at each step should be the same. To solve this we introduce a process that converts variable dimension CNN features to a fixed dimension vector. As we view a path in the tree as a sequence, we model it via an RNN \cite{Graves2009}. Taking converted CNN features as input, the RNN is trained to produce predictions of the target path $y$ in the class tree. 

Formally, a network of $L$ convolution-pooling layers is defined as 
		\begin{equation*}
		\begin{aligned}
		& a_l = f \: (a_{l-1}; \phi_l) \quad \text{for  } l = 1,2, \dots ,L 
		\end{aligned}
		\end{equation*}
where $a_0$ is input image $x$, $\phi_l$ are model parameters at layer $l$, and $f$ is a convolution-pooling function. Note that $ a_l \in \mathbb{R}^{D_l \times W_l \times H_l}$; where $D_l$, $W_l$, and, $H_l$ denote depth, width, and height at the $l^{th}$-layer. The general form of the conversion operation of CNN features, $a_s$, fed to the RNN is defined as
		\begin{equation*}
		\begin{aligned}
  & u_t = \frac{1}{|S_t|} \sum_{s \in S_t} g (a_{s}; \alpha_{s}) \quad \text{for } t = 1,2, \dots ,T
		\end{aligned}
		\end{equation*}
where $S_t \subseteq \{1, \dots, L\} $ is a subset of the CNN layers at each step $t$ of the RNN such that the subsets are ``increasing;'' i.e. for every $1\le n < T$ we have if $i\in S_n, j\in S_{n+1}$, then $i<j$. Also, $g$ is a function of converting CNN outputs into RNN inputs, i.e. $g:\mathbb{R}^{D_s \times W_s \times H_s} \times \mathbb{R}^{\nu_s} \rightarrow \mathbb{R}^{p}$ where $p$ is a dimension of the RNN input at each step and $\alpha_{s} \in \mathbb{R}^{\nu_s}$ are possible trainable model parameters. Conversion methods are discussed in Section \ref{sec:fixCNN}. The RNN takes converted fixed dimension CNN features $u^t \in \mathbb{R}^p$ as inputs, and predicts labels for each layer of the class tree. The RNN is governed by 
		\begin{equation} \label{rnn}
		\begin{aligned}
		& h_t = r_h \: (u_{t}, h_{t-1}; \theta_h) \text{, and } \\
        & o_t = r_o \: (h_{t}; \theta_o) \quad \text{for  } t = 1,2, \dots , T
		\end{aligned}
		\end{equation}
where $h_0$ is an initial hidden state, $r_h$ and $r_o$ are the state transition and output functions, and $\theta_h$ and $\theta_o$ are trainable parameters.

The loss function reads
        \begin{equation*}
		\begin{aligned}        
        \quad \sum_{t=1}^T w_t \: \cdot \: CE(y_t,\text{softmax}(o_t))
		\end{aligned}
        \end{equation*}
where $w_t$ represents weight for level $t$ in $R$ and $CE$ denotes the cross entropy. The aim is for $o_t$ to predict a node in $R$ at level $t$. By definition of RNN all $o_t$ have to have the same dimension. However, the number of classes at each level in $R$ varies. For this reason, we have $o_t \in \mathbb{R}^N$ with $N+1$ being the total number of classes (nodes) in $R$. Label vector $y_t$ is then the one-hot encoding with respect to an $N$-dimensional vector. In inference we employ beam search to find the most likely predicted path in $R$.

We improve the FPL model by applying residual learning to the RNN part of the model. Residual learning for deep networks is introduced by He et al.~\cite{He2015}. We connect the residual arc between input $u_t$ (converted CNN features) and output $o_t$ of the RNN in order to prevent the original CNN features from losing much information while the RNN is trained. Outputs of the RNN with residual learning are calculated by
		\begin{equation*}
		\begin{aligned}
		& o_t^{\text{residual}} = u_t + z \: (o_t; \xi) \quad \text{for  } t = 1,2, \dots, T
		\end{aligned}
		\end{equation*}
where $z$ is a linear mapping function to align dimensions of $u_t$ and $o_t$ with $\xi$ being trainable parameters. In the loss function $o_t$ is then replaced by $o_t^{\text{residual}}$.

\subsection{General tree model: CNN-S2S} \label{sec:general}
In this section, we propose CNN-S2S to fit a general class tree that has target paths with variable lengths from the root to the final class node. In this model, the assumption of a fixed-length of class path in the FPL model is relaxed, and the RNN part of the FPL model is replaced by an S2S. Traditional RNNs are limited to solving problems where input and target sequences are of the same length. S2S introduces an encoder to transform the input sequence to a fixed-dimension representation, and a decoder to process this fixed-length representation to a variable-length sequence \cite{Cho2014}. We propose a model combining a CNN with an S2S that can deal with target label paths of variable lengths. Let now $R$ be a general rooted tree. Label path $y$ is any path from the root to a node (not necessarily a leaf) in $R$. We denote by $|y|$ the number of nodes in $y$. Similar to the FPL model, the general tree model feeds fixed dimension CNN features to S2S, and predicts labels as a vector with the same length as the number of classes in $y$.

Given the converted CNN features $u_t$ as defined in Section \ref{sec:FPL} and the encoder presented by (\ref{rnn}), the decoder reads
		\begin{equation*}
		\begin{aligned}
        & \bar{h}_t = \bar{r}_h \: (\bar{o}_{t}, \bar{h}_{t-1}; \bar{\theta}_{h}), \\
        & \bar{o}_t = \bar{r}_o \: (\bar{h}_{t}; \bar{\theta}_o) 
        \end{aligned}
		\end{equation*}
for  $t = 1,2, \dots , |y|$, and $\bar{h}_0 = h_T$. The loss function defined in the FPL model is used in the general tree model, and  beam search is again used in inference.

\subsection{Conversion operation} \label{sec:fixCNN}

In our models, the conversion operation is the operation of connecting the CNN and the RNN. This operation is needed in order to feed extracted CNN features that have varying dimensions to the RNN that requires the one fixed input dimension. We design conversion operations not only to align related dimensionality, but also to retain much information of the learned CNN representations. 
 
We first describe a linear conversion that converts CNN features directly through the $n$-mode product of a tensor. This conversion is a series of linear transformations to modify dimensionality of the CNN features that requires to train additional model weights. This operator was previously proposed in \cite{Wehrmann2018,Zhu2017}. The linear conversion, $g (a_{s}; \alpha_s)$, is defined as
		\begin{equation*}
		\begin{aligned}
		& g (a_{s};\alpha_s) = \text{vec}( a_{s} \times_{1} U_{s}^{1} \times_{2} U_{s}^{2} \times_{3} U_{s}^{3}) \\
		& s \in S_t, t = 1,2, \dots ,T
		\end{aligned}
		\end{equation*}
where $U_{s}^{1} \in \mathbb{R}^{ m \times D_s}$, $U_{s}^{2} \in \mathbb{R}^{ n \times W_s},$ and $U_{s}^{3} \in \mathbb{R}^{ v \times H_s}$ are trainable parameter matrices at CNN layer $s \in S_t$ ($\alpha_s = (U_s^1,U_s^2,U_s^3) $). Here $\times_k$ is the $k$-mode product of a tensor by a matrix and $\text{vec}$ is a flattening operation. In this conversion, a desired RNN or S2S input dimension, $p$, is determined by $m \cdot n \cdot v = p $. 

We also propose conversion methods by using convolutional and pooling operations. Convolutional conversion retains spatial information of the CNN features and aligns related dimensions efficiently. However, we have additional trainable model weights similar to the linear conversion. The convolutional conversion, $g (a_{s}; \alpha_s)$, is defined as
		\begin{equation*}
		\begin{aligned}
		& g (a_{s};\alpha_s) = \text{vec}( \text{Conv} (a_{s};\alpha_s)) \: s \in S_t, t = 1,2, \dots ,T
		\end{aligned}
		\end{equation*}
where $\alpha_s$ are trainable parameters for convolution operation Conv. Note that the filter size, stride, and depth of Conv have to be selected in such a way that the resulting vector is in $\mathbb{R}^p$. The details are provided in the appendix.

The pooling conversion does not require to train additional model weights, at the same time it keeps spatial information of the original CNN features. The pooling conversion, $g (a_{s};\alpha_s)$, is defined as
		\begin{equation*}
		\begin{aligned}
		& g (a_{s};\alpha_s) = \text{vec}( \text{Pool} (a_{s} ) \times_{1} U_{s}^{1} \times_{2} U_{s}^{2} \times_{3} U_{s}^{3}) \\ 
		& s \in S_t, t = 1,2, \dots ,T
		\end{aligned}
		\end{equation*}
where Pool is the pooling operation. The details are provided in the appendix. 



\section{Computational Study} \label{sec:exp}

In this section we present two cases: one based on a proprietary dataset and the other one based on a public dataset. The models have been implemented using Tensorflow. A single GPU card has been used in every run. The code is available at \url{http://github.com/to_be_released_after_acceptance}. In the experiments, we compare our models to state-of-the-art CNNs: CNN architectures by Visual Geometry Group (VGG) \cite{Simonyan2014}, Residual neural networks (Res) \cite{He2015}, Squeeze-and-Excitation networks (SE) \cite{Hu2018}, and Convolutional Block Attention Module (CBAM) \cite{Woo2018}. For the RNN and S2S parts of our networks a bidirectional RNN with LSTM cells is applied.

\subsection{Real world dataset} \label{real:trandeval}

We have conducted experiments on a real world proprietary dataset containing approximately 180,000 images. We selected validation and test sets with approximately 36,000 images each. The classes have mainly Has-A hierarchical relationships with 18 classes and the tree of depth four in the FPL tree, and 15 classes (nodes in the tree) and the tree of maximal depth four in the general tree. The general tree setting has target paths with lengths ranging from two to four.


Preprocessing of raw data and hyperparameters are determined based on \cite{He2015,Simonyan2014}. Original images are resized with its shorter side sampled in [256, 512] and then cropped to 224$\times$224. Hyperparameters are set as follows: batch size is set to 32, input dimensions of the RNN converted from CNN features range from 512 to 4,096, the dimensionality of the RNN hidden states range from 512 to 1,024, the FPL model has three RNN layers, and the general tree model has one S2S layer. The CNN part is initialized with the weights trained on ImageNet. Orthogonal random initialization is adapted for the RNN and S2S weights. Due to low memory requirements, we use the pooling conversion for the real dataset since it it the only option on this dataset. The conversion operations are composed later on the public dataset where it is established that pooling is best.

We apply an alternating training scheme for the CNN and RNN parts; i.e. we update one while keeping the other frozen and flip in the beginning of training, and then in a later phase unfreeze the entire network. Alternating prevents divergence during training of the CNN and the RNN part as each has a distinct purpose; the CNN learns hierarchical features of images, and the RNN learns hierarchical trees of categories. In addition, the quality of weight initializations is uneven between the CNN and the RNN as the CNN starts with high-quality pretrained weights from a large-scale dataset, ImageNet, while the RNN starts with random weights. For this reason, in our training we unfreeze the RNN first in the alternating scheme. These settings are applied to VGG-16 and Res-50.

To evaluate performance of our models, two metrics are compared. For path accuracy we count a prediction correct if the entire predicted path matches all of the labels in the ground truth while for node accuracy we count how many nodes in the target path are correct in the predicted path. Node accuracy captures how accurately predictions fit the ground truth at different levels. Remark 1: Because as simple CNN classifiers, VGG-16 and Res-50, can predict only the final node of a path and for compatibility of the metric we imply that if the final node is correctly predicted, all its predecessors are correctly predicted as well.


In Tables \ref{BT-table} and \ref{GT-table} accuracies on the test set are presented. CNN models are denoted by Res-50 and VGG-16. Our FPL and general tree models are denoted by (CNN structure)-(RNN structure)-(alternating training scheme)-(residual learning). Training takes around six days for 30 epochs. 

\begin{table} 
\begin{center}
\begin{tabular}{|l|c|c|}
\hline
\multirow{2}{*}{Method} & \multicolumn{2}{c|}{Accuracy (\%)} \\ \cline{2-3}
                        & Path   & Node  \\
\hline
VGG-16            & 64.0 & 71.3 \\
VGG-RNN           & 59.9 & 77.3 \\
VGG-RNN-Alt       & 61.7 & 78.1 \\
VGG-RNN-Alt-Resi  & \textbf{65.3} & \textbf{80.5} \\
\hline
Res-50               & 62.1 & 68.8 \\
Res-RNN              & 63.4 & 78.9 \\
Res-RNN-Alt          & 62.4 & 77.0 \\
Res-RNN-Alt-Resi     & \textbf{64.0}  & \textbf{78.4} \\
\hline
\end{tabular}
\end{center}
\caption{Test accuracies of the FPL model on real data}
\label{BT-table}
\end{table}

\textbf{FPL model}: Table \ref{BT-table} presents the experimental results of the FPL model. Residual variants with alternating training perform best on both Res and VGG. This shows that both CNN and RNN of our model successfully play their specific roles; the CNN learns hierarchical features of images, and the RNN correctly predicts target paths. Furthermore, residual arcs and alternating training help improving test accuracy. Our models with Res perform better than Res-50 on both path and node accuracies. However, for VGG our non-residual variants VGG-RNN and VGG-RNN-Alt perform worse on path accuracy than VGG-16 even though our models show higher node accuracy. This can be interpreted as our models solving more difficult problems than CNN regarding path accuracy (see Remark 1). Explicitly predicting all nodes along a path in a tree is more difficult than the path correctness implicitly assumed as soon as only the final node is correctly predicted.

\begin{table} 
\begin{center}
\begin{tabular}{|l|c|c|}
\hline
\multirow{2}{*}{Method} & \multicolumn{2}{c|}{Accuracy (\%)} \\ \cline{2-3}
                        & Path   & Node  \\
\hline
VGG-16           & 72.8 & 87.2 \\
VGG-S2S          & \textbf{73.7} & \textbf{88.1} \\
VGG-S2S-Alt      & 72.3 & 87.2 \\
\hline
Res-50        & 71.5 & 86.4 \\
Res-S2S       & \textbf{74.1} & \textbf{88.0} \\
Res-S2S-Alt   & 68.3 & 85.0 \\
\hline
\end{tabular}
\end{center}
\caption{Test accuracies of the general tree model on real data}
\label{GT-table}
\end{table}

\textbf{General tree model}: Table \ref{GT-table} shows experimental results of the general tree model. Our model without alternating training performs better on both path and node accuracies than CNNs. This proves that our models successfully extract hierarchical features of images and learns a label path with variable lengths of target paths. However, alternating training in the experiments did not help to improve performance of our models. This can be interpreted as S2S with orthogonal initialization being good enough to avoid diverging in training.

\subsection{Public dataset: Open Images}

Open Images V4 is a public dataset of 9 million images annotated with image-level labels, object bounding boxes and visual relationships \cite{OpenImages2}. We use a subset of the original dataset for hierarchical classification.
The subset contains approximately 950,000 images with 2.4 million labels for training and 36,000 images with 127,000 labels for test; it is a multi-label dataset. Since the labels reside in different levels of the original class hierarchy, we build a class tree of depth four by concatenating subtrees of the original hierarchy. There are 30 classes in the tree with the classes having mainly Has-A relationships. Figure \ref{hc-oi-tree} presents a subtree of the tree with the full tree presented in the appendix. We follow the same preprocessing steps of raw images and model architectures as those used in Section \ref{real:trandeval}. The dataset is available at \url{http://link_to_be_added_after_acceptance}.

\begin{figure}
\begin{center}
\includegraphics[width=0.7\textwidth]{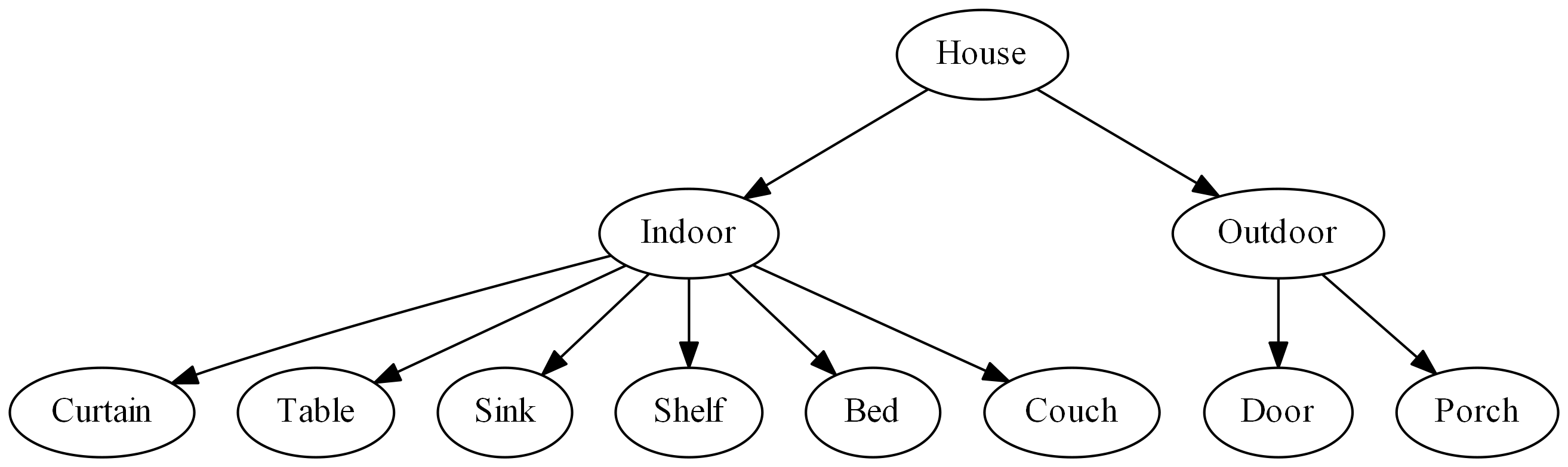}
\end{center}
   \caption{A part of the tree of Open Images}
\label{hc-oi-tree}
\end{figure}

To evaluate multi-label classification models, scores computed by precision and recall are typically considered such as the F1 score and area under precision-recall curve \cite{Bi2011,Vens2008,Zhang2014}. In this study, we select the F1 score as our evaluation metric that computes the harmonic average of the precision and recall.

To solve the multi-label classification problem which in our case corresponds to multiple paths in the tree for a single sample, we select the sigmoid function at the output layer of CNNs and our models rather than the softmax function. The predictions are selected as those with the logit value above a threshold. 
A path is selected if all logits of the nodes in the path are above the threshold. 
The threshold is selected so as to maximize the F1 score on the validation dataset. It is then used on the test dataset.
In the same way as in the real data experiments, we calculate the path and node F1 scores. To provide more reliable results, we compute the means of the test F1 scores and their standard deviations for each model averaged over 3 random runs. Each training takes around four days for 15 epochs.

\begin{table} 
\begin{center}
\begin{threeparttable}
\begin{tabular}{|l|c|c|}
\hline
\multirow{2}{*}{Method} & \multicolumn{2}{c|}{Mean F1 score and Sd. (\%)} \\ \cline{2-3}
                 & Path     & Node  \\
\hline
VGG-16              & 68.62 (0.09)& 72.29 (0.09)\\
VGG-S2S-Linear      & \textbf{71.49 (0.10)}& \textbf{74.84 (0.12)}\\
VGG-S2S-Alt-Linear  & 70.76 (0.06)& 73.97 (0.15)\\
VGG-S2S-Conv        & 68.61 (0.41)& 71.62 (0.51)\\
VGG-S2S-Alt-Conv    & 70.70 (0.14)& 73.95 (0.06)\\
VGG-S2S-Pool        & 71.39 (0.16)& 74.76 (0.18)\\
VGG-S2S-Alt-Pool    & \textbf{71.49 (0.10)}& \textbf{74.84 (0.07)}\\
$\text{VGG-S2S-Pool-}\mathcal{L}$       & \textbf{71.93 (0.09)}& \textbf{75.29 (0.07)}\\
$\text{VGG-S2S-Alt-Pool-}\mathcal{L}$   & 71.26 (0.13)& 74.58 (0.17)\\
\hline
Res-50             & 71.15 (0.04)& 74.47 (0.07)\\
Res-S2S-Linear\tnote{*}     & \textbf{72.14 (0.14)}& \textbf{75.46 (0.16)}\\
Res-S2S-Alt-Linear & 70.78 (0.09)& 73.80 (0.14)\\
Res-S2S-Conv       & 70.42 (0.39)& 73.55 (0.44)\\
Res-S2S-Alt-Conv   & 70.75 (0.15)& 73.98 (0.18)\\
Res-S2S-Pool       & \textbf{71.98 (0.06)}& \textbf{75.40 (0.07)}\\
Res-S2S-Alt-Pool   & 71.70 (0.06)& 74.97 (0.09)\\
$\text{Res-S2S-Pool-}\mathcal{L}$\tnote{*}     & \textbf{72.05 (0.06)}& \textbf{75.43 (0.03)}\\
$\text{Res-S2S-Alt-Pool-}\mathcal{L}$  & 71.66 (0.04)& 74.89 (0.05)\\
\hline
CBAM-Res-50             & 71.17 (0.09)& 74.42 (0.04)\\
CBAM-Res-S2S-Linear     & \textbf{71.71 (0.19)}& \textbf{74.98 (0.19)}\\
CBAM-Res-S2S-Alt-Linear & 66.44 (0.43)& 68.84 (0.53)\\
CBAM-Res-S2S-Conv       & 70.24 (0.07)& 73.43 (0.05)\\
CBAM-Res-S2S-Alt-Conv   & 72.88 (0.09)& 69.82 (0.07)\\
CBAM-Res-S2S-Pool       & \textbf{71.53 (0.12)}& \textbf{74.82 (0.10)}\\
CBAM-Res-S2S-Alt-Pool   & 71.29 (0.18)& 74.55 (0.11)\\
CBAM-$\text{Res-S2S-Pool-}\mathcal{L}$      & \textbf{71.57 (0.07)}& \textbf{74.89 (0.07)}\\
CBAM-$\text{Res-S2S-Alt-Pool-}\mathcal{L}$  & 71.32 (0.05)& 74.55 (0.01)\\                 
\hline
SE-Res-50             & 71.22 (0.10)& 74.49 (0.08)\\
SE-Res-S2S-Linear\tnote{*}     & \textbf{72.05 (0.04)}& \textbf{75.33 (0.08)}\\
SE-Res-S2S-Alt-Linear & 69.52 (0.11)& 72.45 (0.14)\\
SE-Res-S2S-Conv       & 70.49 (0.15)& 73.65 (0.14)\\
SE-Res-S2S-Alt-Conv   & 70.35 (0.09)& 73.51 (0.06)\\
SE-Res-S2S-Pool       & \textbf{71.79 (0.15)}& \textbf{75.12 (0.16)}\\
SE-Res-S2S-Alt-Pool   & 71.42 (0.05)& 74.73 (0.04)\\
SE-$\text{Res-S2S-Pool-}\mathcal{L}$      & \textbf{71.75 (0.09)}& \textbf{75.07 (0.04)}\\
SE-$\text{Res-S2S-Alt-Pool-}\mathcal{L}$ & 71.07 (0.13)& 74.27 (0.18)\\
\hline
\end{tabular}
\begin{tablenotes}
\item [$\mathcal{L}$ denotes models with larger S2S than others.]
\item [$*$ denotes top three performers in the table.]
\end{tablenotes}
\end{threeparttable}
\end{center}
 \caption{Test F1 scores of general tree on Open Images}
\label{GT-OI-table}
\end{table}

Table \ref{GT-OI-table} presents the test F1 scores of the general tree model (FPL does not apply here since the paths have different lengths). VGG-16, Res-50, CBAM-Res-50, and SE-Res-50 are CNN models. Our general tree models are denoted by (CNN structure)-(RNN structure)-(alternating training scheme)-(conversion methods). For our models with Linear, Conv, and Pool, we use 256 and 512 for input and hidden state dimensions of S2S, respectively. Under this setting all three conversion operations can be executed without a memory problem. For our models with Pool, we also use larger dimensions of 2048 and 1024 for input and hidden state of S2S, respectively, since the pooling conversion has lower memory requirements than others. We stress in bold the best three models in the table.
We find that SE-Res-50 performs best among all CNN models.
All of our models based on different CNN architectures such as VGG, Res, SE, and CBAM perform better on path and node F1 scores than standard CNN models. The largest improvement over CNN models is made by our model $\text{VGG-S2S-Pool-}\mathcal{L}$, and the highest F1 score is achieved by Res-S2S-Linear.
This shows that our models fit the multi-label class tree better by extracting CNN features from images and predicting paths with variable lengths by S2S. 
Alternating training does not work for S2S in many cases even though our model with alternating is one of the best three performers in VGG.
As pointed out in real data experiments, the effect of alternating training can be overshadowed by other factors such as weight initialization. In almost all cases, pooling conversion performs better than other conversion methods. This is because pooling at conversion keeps the spatial information of the trained CNN features. 
However, convolutional conversion works only in VGG when it is combined with alternating training. As this requires additional trainable weights, it makes our models harder to train and consumes additional memory. We find that its usage may be limited to smaller networks. 
We also find that models with larger S2S vector sizes perform better than smaller models when alternating training is not used. 
Our models feed CNN features to S2S, and so the input to S2S is a representation of images. Since larger S2S models can extract more information from images from trained CNN features than smaller models, the larger models are expected to achieve better results. 
This perspective is not true for alternating training even though larger models without alternating perform the best among models with pooling conversion. 



\section{Conclusion} \label{sec:conclusion}

In this work, we develop a new structure of deep neural networks for hierarchical classification of images. Combining CNN as a feature learner with RNN or S2S as a sequence classifier, the proposed models can predict a target path in a hierarchical tree of classes. By means of capturing hierarchical representations, the proposed models take features from different CNN layers, and feed them to RNN or S2S. Depending on the class tree structure two models are suggested; the FPL model (CNN-RNN) and the general tree model (CNN-S2S) for a fixed- and variable-length target paths. To expedite training and improve generalization of the model, we also suggest a CNN-RNN variation that adds residual arcs to the RNN part. To examine the performance of our models, we conduct experiments on a proprietary and a public dataset of images. Experimental results show that our models perform better than state-of-the-art CNNs, VGG, Res, SE, and CBAM. For CNN-RNN, our models with residual arcs perform best in predicting fixed length paths for both CNN networks. For CNN-S2S, our models without alternating training perform the best in almost all cases. For this reason we recommend not to use alternating training in the S2S case. Considering GPU memory is limited in practice, we recommend to use pooling conversion as it performs well and has low memory requirements for training. 

\medskip
\small
\bibliography{main}

\appendix

\section{Deciding dimensions for convolutional conversion} \label{con:conv}  
Let $F_{conv_s}$, $K_{conv_s}$, $G_{conv_s}$, and $Z_{conv_s}$ denote the filter size, number of filters, stride, and zero padding for convolution Conv for $s \in S_t$, respectively. Also, let $D_{conv_s}$, $W_{conv_s}$, and $H_{conv_s}$ denote the depth, width, and height of output after the convolution operation. In this conversion, the desired RNN or S2S input dimension, $p$, is determined by $ D_{conv_s} \cdot W_{conv_s} \cdot H_{conv_s} = p$. The formulas for convolution lead to $D_{conv_s} = K_{conv_s}; W_{conv_s} = \frac{W_s - F_{conv_s} + 2 Z_{conv_s}}{G_{conv_s}}+1;$ and $H_{conv_s} = \frac{H_{s} - F_{conv_s} + 2 Z_{conv_s}}{G_{conv_s}}+1$. We assume that $H_{s}$ and $W_{s}$ are the same, i.e. the feature map is shaped in a square. \\
1) $W_{s} = H_{s} = F_{conv_s}$: We have 
		\begin{equation*}
		\begin{aligned}
		& p = D_{conv_s} \cdot W_{conv_s} \cdot H_{conv_s} \\
        & \: \: \: = K_{conv_s} \cdot (\frac{W_s - F_{conv_s} + 2 Z_{conv_s}}{G_{conv_s}}+1) \cdot (\frac{H_{s} - F_{conv_s} + 2 Z_{conv_s}}{G_{conv_s}}+1),
		\end{aligned}
		\end{equation*}
and thus $K_{conv_s} = p, Z_{conv_s} = 0$,  and  $G_{conv_s} \in \mathbb{Z}^{+}$. \\
2) $ W_{s} = H_{s} > F_{conv_s}$, $ F_{conv_s} = G_{conv_s} $: We have
		\begin{equation*}
		\begin{aligned}
		& p = D_{conv_s} \cdot W_{conv_s} \cdot H_{conv_s} \\
        & \quad = K_{conv_s} \cdot (\frac{W_s - F_{conv_s} + 2 Z_{conv_s}}{G_{conv_s}}+1) \cdot (\frac{H_{s} - F_{conv_s} + 2 Z_{conv_s}}{G_{conv_s}}+1) \\
        & \quad = K_{conv_s} \cdot (\frac{W_s - F_{conv_s} + 2 Z_{conv_s}}{F_{conv_s}}+1)^2 \\
        & \quad = K_{conv_s} \cdot (\frac{W_s + 2 Z_{conv_s}}{F_{conv_s}})^2. \\
		\end{aligned}
		\end{equation*}
This leads to $F_{conv_s} \in  [1, W_{s}), Z_{conv_s}$ such that $(\frac{W_s + 2 Z_{conv_s}}{F_{conv_s}})^2 \in \mathbb{Z}^{+}$, and $K_{conv_s}$ such that $\frac{p}{(\frac{W_s + 2 Z_{conv_s}}{F_{conv_s}})^2} \in \mathbb{Z}^{+}$. \\
3) $ W_{s} = H_{s} > F_{conv_s}, F_{conv_s} \neq G_{conv_s}$, $G_{conv_s} = 1 $: We have
		\begin{equation*}
		\begin{aligned}
		& p = D_{conv_s} \cdot W_{conv_s} \cdot H_{conv_s} \\
        & \quad = K_{conv_s} \cdot (\frac{W_s - F_{conv_s} + 2 Z_{conv_s}}{G_{conv_s}}+1) \cdot (\frac{H_{s} - F_{conv_s} + 2 Z_{conv_s}}{G_{conv_s}}+1) \\
        & \quad = K_{conv_s} \cdot (\frac{W_s - F_{conv_s} + 2 Z_{conv_s}}{G_{conv_s}}+1)^2 \\
        & \quad = K_{conv_s} \cdot (W_s - F_{conv_s} + 2 Z_{conv_s}+1)^2,
		\end{aligned}
		\end{equation*}
and thus $F_{conv_s} \in  [1, W_{s}), Z_{conv_s} = 0,$ and $K_{conv_s}$ such that $\frac{p}{(W_s - F_{conv_s} + 1)^2} \in \mathbb{Z}^{+}$.\\
4) $W_{s} = H_{s} > F_{conv_s}, F_{conv_s} \neq G_{conv_s}$, $G_{conv_s} \in (1, F_{conv_s}) $: We have
		\begin{equation*}
		\begin{aligned}
		& p = D_{conv_s} \cdot W_{conv_s} \cdot H_{conv_s} \\
        & \quad = K_{conv_s} \cdot (\frac{W_s - F_{conv_s} + 2 Z_{conv_s}}{G_{conv_s}}+1) \cdot (\frac{H_{s} - F_{conv_s} + 2 Z_{conv_s}}{G_{conv_s}}+1) \\
        & \quad = K_{conv_s} \cdot (\frac{W_s - F_{conv_s} + 2 Z_{conv_s}}{G_{conv_s}}+1)^2, \\
		\end{aligned}
		\end{equation*}
implying that $F_{conv_s} \in  [1, W_{s}), G_{conv_s} \in (1, F_{conv_s}), Z_{conv_s}$ such that $(\frac{W_s - F_{conv_s} + 2 Z_{conv_s}}{G_{conv_s}}+1)^2 \in \mathbb{Z}^{+}$, and $K_{conv_s}$ such that $\frac{p}{(\frac{W_s - F_{conv_s} + 2 Z_{conv_s}}{G_{conv_s}}+1)^2} \in \mathbb{Z}^{+}$.

\section{Deciding dimensions for pooling conversation} \label{con:pool}
A desired RNN input dimension, $p$, is determined by satisfying $ D_{pool_s} \cdot W_{pool_s} \cdot H_{pool_s} = p$. The pooling operation leads to $D_{pool_s} = D_s ; W_{pool_s} = \frac{W_s - F_{pool_s}}{G_{pool_s}}+1;$ and $H_{pool_s} = \frac{H_s - F_{pool_s}}{G_{pool_s}}+1$. We assume that $H_{s}$ and $W_{s}$ are the same, i.e. the feature map is shaped in a square, and $F_{pool_s}$ and $G_{pool_s}$ are the same, i.e. disjoint pooling. \\
1) $H_{s} = W_{s}$, $F_{pool_s} = G_{pool_s}$: We have
		\begin{equation*}
		\begin{aligned}
		& p = D_{pool_s} \cdot W_{pool_s} \cdot H_{pool_s} \\
        & \quad = D_{s} \cdot (\frac{W_s - F_{pool_s}}{G_{pool_s}}+1) \cdot (\frac{H_s - F_{pool_s}}{G_{pool_s}}+1) \\
        & \quad = D_{s} \cdot (\frac{b_s - a_{pool_s}}{a_{pool_s}}+1)^2 \\
        & \quad = D_{s} \cdot (\frac{b_s}{a_{pool_s}})^2, 
		\end{aligned}
		\end{equation*}
and thus $a_{pool_s} = b_s \cdot \sqrt{\frac{D_s}{p}} = F_{pool_s} = G_{pool_s}$.\\
2) $H_{s} = W_{s}$, $F_{pool_s} \neq G_{pool_s}$: We have
		\begin{equation*}
		\begin{aligned}
		& p = D_{pool_s} \cdot W_{pool_s} \cdot H_{pool_s}\\
        & \quad = D_{s} \cdot (\frac{W_s - F_{pool_s}}{G_{pool_s}}+1) \cdot (\frac{H_s-F_{pool_s}}{G_{pool_s}}+1) \\
        & \quad = D_{s} \cdot (\frac{W_s - F_{pool_s}}{G_{pool_s}}+1)^2, 
		\end{aligned}
		\end{equation*}
leading to $F_{conv_s} \in [1,W_{s})$, and $G_{conv_s} \in (1,F_{conv_s})$.

\section{Description of the general tree of Open Images}

\begin{figure*}
\begin{center}
\includegraphics[width=1.0\textwidth]{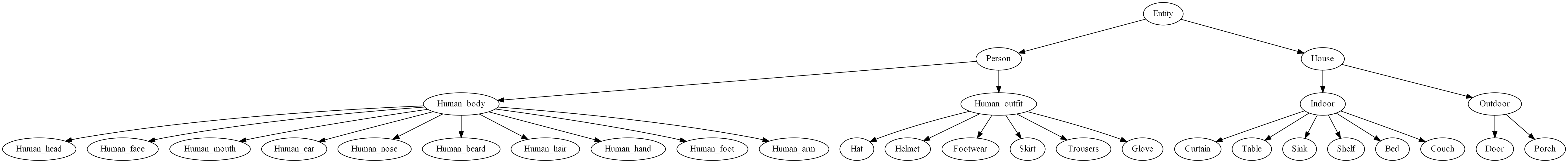}
\end{center}
   \caption{The full general tree of Open Images}
\label{hc-oi-tree-full}
\end{figure*}

Open Images provides a semantic hierarchy that consists of 600 object classes \cite{OpenImages2}. We first take subtrees of classes that have mainly Has-A relationships such as `Human body'-`Human foot' and `Human outfit'-`Hat,' and then concatenate the subtrees to create the final class tree. The tree is of depth four with 30 class nodes. Figure \ref{hc-oi-tree-full} presents the whole tree.

\end{document}